%% file: main.tex
\documentclass[10pt,twocolumn]{article}
\pdfoutput=1
\usepackage{spconf,amsmath,graphicx,subcaption,booktabs}

\name{\begin{tabular}{c}Hmrishav Bandyopadhyay$^{{\dagger}}$ Shuvayan Ghosh Dastidar$^{{\dagger}}$ Bisakh Mondal$^{{\dagger}}$ \\
		Biplab Banerjee$^{{\star}}$ Nibaran Das$^{{\dagger}}$\end{tabular}}

\address{$^{{\dagger}}$ Jadavpur University, Kolkata, India \hspace{0.5em} $^{{\star}}$Indian Institute of Technology, Bombay, India}

\begin{document}

\title{A distillation based approach for the diagnosis of diseases}

\maketitle
\begin{abstract}


Presently, Covid-19 is a serious threat to the world at large. Efforts are being made to reduce disease screening times and in the development of a vaccine to resist this disease, even as thousands succumb to it everyday. We propose a novel method of automated screening of diseases like Covid-19 and pneumonia from Chest X-Ray images with the help of Computer Vision. Unlike computer vision classification algorithms which come with heavy computational costs, we propose a knowledge distillation based approach which allows us to bring down the model depth, while preserving the accuracy. We make use of an augmentation of the standard distillation module with an auxiliary intermediate assistant network that aids in the continuity of the flow of information. Following this approach, we are able to build an extremely light student network, consisting of just 3 convolutional blocks without any compromise on accuracy. We thus propose a method of classification of diseases which can not only lead to faster screening, but can also operate seamlessly on low-end devices.


\end{abstract}
\section{Introduction}
\label{sec:intro}

Detection of diseases from Chest X-ray Radiography (CXR) images can be done by radiologists who can spot abnormalities in bones and blood vessels accurately. This, however, requires considerably expertise and experience, which not only raises the costs, but results in an incorrect diagnosis when performed by inexperienced individuals.


Computer aided diagnosis of CXRs is not only super-fast, but can also bring down the cost of detection of several diseases. With the advent of several computer vision algorithms, we have seen a multitude of ideas for the detection and classification of diseases from chest X-ray images with particular focus on pneumonia and tuberculosis ~\cite{Wang_2017_CVPR,asnaoui2020automated,lakhani2017deep,hwang2019development}. Neural networks for medical diagnosis however, have been increasing in depth and parameter count over the years to meet the increasing demands in performance, resulting in slower predictions, thereby defeating the main motive of automated screening in the first place.

To curb the excessive parameter count in deep learning methods for diagnosis of diseases, we propose the use of the concept of knowledge distillation for the transfer of knowledge from very deep teacher networks to smaller student networks which can give predictions with accuracy at par with the teacher network. To minimize the distance between feature representations of the very deep teacher network and the extremely light student network, an assistant model of intermediate depth ~\cite{mirzadeh2019improved} is made use of.


Thus our model overview consists of a teacher network, an assistant, and a student network. The teacher network has the backbone of an Efficient-Net module~\cite{tan2020efficientnet} and is used to generate soft-labels. These soft labels represent the confidence scores of the teacher network's predictions and help us understand the mapping of the teacher's feature vectors to the labels. 
The soft labels generated by the teacher network are used as ground truth by the assistant, which in turn generates more such labels for the student model to train on. We summarise our primary contributions as:
\begin{enumerate}
    \item Extending knowledge distillation to the field of medical imaging. Using these methods on multi-label datasets, specifically for easy and fast detection of Covid-19.
    \item Development of extremely light-weight models which can detect diseases in a fraction of a second.
    \item Incorporation of an intermediate assistant network for continual flow of information from the teacher to the student network.
\end{enumerate}
\section{Related Work}
Classification and localization of diseases with the help of Chest X-Rays (CXRs) has been recently a primary topic of exploration for Computer Vision and Medical Imaging researchers. The publication of the ChestX-Ray14 (NIH) dataset has contributed to a series of works relating to the classification and localization of diseases. Wang \emph{et al.} evaluated several state-of-the-art convolutional neural network
architectures in \cite{Wang_2017_CVPR} such as AlexNet, Resnet, VGGNet, GoogLeNet to predict the presence of different diseases in CXRs, reporting the Area Under the ROC curve (AUC).  Rajpurkar \emph{et al.}~\cite{rajpurkar2017chexnet} demonstrated that a common DenseNet architecture~\cite{huang2018densely} can surpass the accuracy of radiologists in detecting pneumonia. In addition
to a DenseNet, Yao \emph{et al.}~\cite{8727965} implemented a Long-short Term
Memory (LSTM) model to exploit dependencies between the
abnormalities in CXRs. Bar \emph{et al.} presented an early examination of the strength of deep learning approaches for pathology detection in ~\cite{7163871}. Li \emph{et al.}~\cite{Liu_2019_ICCV} proposed to unify the training of image-level and box-level labels in one framework with a customized MIL(Multiple Instance learning) loss where disease classification is performed on a grid over the image.

We find several works which have applied an attention mechanism for chest X-ray analysis. For example, Guan \emph{et al.}~\cite{guan2018diagnose} designed an attention guided two-branch network for thorax disease classification, which helps to amplify the high activation regions. Cai \emph{et al.}~\cite{cai2018iterative} presented an
attention mining strategy to improve the model’s sensitivity or saliency to disease patterns. Rubin \emph{et al.}~\cite{rubin2018large} designed a Dual-Network to extract the image
information of both frontal and lateral views.
 Irvin \emph{et al.}~\cite{irvin2019chexpert} trained
on a dataset where the ground truth consists of an additional
uncertainty class. Different approaches were applied during
training to increase the performance with the uncertainty
information.
\par The recent advent of the COVID-19 pandemic had led to a number of works focusing on screening the disease with the help of CXRs, fed into deep neural networks. We find an example, in ~\cite{das2020truncated}, where a CNN was applied with a backbone of the Inception network to detect COVID-19 disease from tomography scans. Moreover, in ~\cite{taresh2020transfer}, a CNN
architecture called COVID-Net based on transfer learning
was applied to classify the CXR images into three classes:
normal, COVID-19 and pneumonia. All of these Neural networks used for disease classification, however contain a huge number of parameters, making them ineffective for fast predictions. A method of reducing the depth of a neural network while retaining its accuracy was first proposed by Hinton \emph{et al.} in ~\cite{hinton2015distilling}. In here,  the concept of knowledge distillation was introduced through experiments on Mnist and datasets on speech recognition, where they were able to achieve satisfactory results. Following this, there were series of papers employing the concept of knowledge distillation. Papers which built up from this idea include a work by Jang \emph{et al.} in ~\cite{cho2019efficacy} where a thorough evaluation of the efficacy of knowledge distillation and its dependence on student and teacher architectures was presented.
Wang \emph{et al.} ~\cite{Wang_2021} provides a comprehensive survey of teacher-student frameworks in the domain of computer vision. Xiaodong \emph{et al.} also performed experiments of knowledge distillation on multi-task deep neural network for natural language processing in ~\cite{liu2019improving}. Mirzadeh et. al ~\cite{mirzadeh2019improved} proposed an assistant model for bridging the gap between the parameter space of the teacher model and the student model.

\section{Our Framework}


Inspired by ~\cite{hinton2015distilling}, we apply the method of knowledge distillation to generate feature maps $ f_T $ from $ n_s $ (total training samples) labeled data points $ \{ (x_i, y_i) \}_{i=1}^{n_s} $ for classification. As in standard distillation techniques, we first train the teacher network to capture features $ f_T $ it finds in images with the help of CNNs. To further improve the stability of the student network, we introduce an assistant network  network which is trained to replicate the feature extraction methods of the teacher on a lower parameter set with the help of feature maps and soft labels $y_i^s$ generated by the teacher.
After the teacher and the assistant networks have been trained, we train the student network with the soft labels generated from the assistant network and the feature maps it creates. We make use of various distance metrics to compare between feature maps as loss functions for training the model.

\subsection{The Teacher Network}
The teacher network $ C_T $, is a DCNN with the backbone of an Efficient-Net~\cite{tan2020efficientnet} module. The entire model architecture has been summarised in Fig. \ref{fig:teacher}.

\input{fig_eff.tex}


We use the EfficientNet model as the primary model that extracts information from the CXRs in the form of feature maps and produces soft labels by attempting to classify the feature maps into hard-labels. Particularly, we use the B6 and B7 models as in Fig. \ref{fig:teacher} from the EfficientNet group and perform an ensemble of them to obtain the final results. To train the network on the multi-class classification task, we make use of Binary Cross Entropy per class as a loss function. The BCE Loss function can be mathematically expressed as : 
\begin{equation}
L_{C_T}=\sum_{i=0}^{C} [-(y_{i}log(\hat{y_{i}})) -(1-y_{i})(log(1-\hat{y_{i}}))]
\end{equation}
where $y_{i}$ represents the per class ground truth value and  $ \hat{y_{i}} $ represents the model outputs.
To help the student network $ C_S$ get a holistic idea of the labels, we use a temperature to regulate the labels generated by the teacher. The higher we set the value of the temperature, the 'softer' the labels go. The temperature regulated Softmax can be expressed mathematically as :
\begin{equation}
y_{i}^s=\frac{\exp{(\hat{y_{i}}/T)}}{{\sum_{j}^{C}\exp{(\hat{y_{j}}/T)}}}
\end{equation}
where $i$ is the class label for a particular training data, $C$ is the total number of classes, and $T$ is the value of the temperature as a constant.

\subsection{The Assistant Network}

For the assistant network $ C_A $, we use the Densenet 121 backbone. The DenseNet network is preferred as it can attain superior feature propagation in a very low parameter space. While the assistant network has fewer parameters than the teacher network, its parameters are much high compared to the student. Thus, using it to learn from the teacher network helps us to reduce the loss of information we undergo when we distill the student directly from the teacher. 

A diagrammatic representation of the assistant can be estimated from Fig. \ref{fig:student}-A. We train the assistant network with the help of BCE loss against the original hard labels and soft labels generated by the teacher. We also compare its feature maps with those of the teacher using KL divergence as a distance metric. The mathematical form of our loss function for the assistant can be expressed as:
\begin{equation}
\begin{split}
L_{C_A}=\sum_{i=0}^{C} [&-(y_{i}log(\hat{y_{i}}))  - (1-y_{i})(log(1-\hat{y_{i}}))] \\
& + \lambda_{1}\mathcal{E}_{x\sim P}[\log\frac{f_T}{f_A}] + L_{soft}
\end{split}
\end{equation}
where $y_{i}$ and  $\hat{y_{i}}$ are usual notations and $f_T$ \& $f_A$ are intermediate feature maps obtained from teacher and assistant respectively. $L_{soft}$ is the BCE loss between the soft labels generated from the teacher and the assistant's predictions.

\subsection{The Student Network}


The student $C_S$ is a lightweight network as in Fig. \ref{fig:student} that consists of 3 consecutive convolutional layers. The primary convolutional layer makes use of 64 filters and has a kernel size of 5x5. The next two layers of the student contain 3x3 kernels with 64 and 128 filters respectively. Wasserstein distance is used as a loss function and is computed between the feature maps at the output of each convolutional layers of the student and feature maps at different levels of the teacher network to help maintain a correlation between them. This ensures that the student learns along the lines of the teacher along with helping in faster convergence. Classification from the feature space is done with the help of linear layers and the loss is computed with the BCE Loss function. In here, temperature regulated soft labels and original hard labels are used as the ground truths. A diagrammatic representation of the student has been given in Fig. \ref{fig:student}-B. The effective total loss function is:
\begin{equation}
\begin{split}
L_{C_S}=\sum_{i=0}^{C} [&-(y_{i}log(\hat{y_{i}})) - (1-y_{i})(log(1-\hat{y_{i}}))] \\
& +\lambda_{2} (\inf E[d(f_S,f_A)^{p}])^{1/p} + L_{soft}
\end{split}
\end{equation}

where $y_{i}$ and  $\hat{y_{i}}$ are usual notations and $ E[Z] $ denotes the expected value of a random variable $ Z $ and the infimum is taken over all joint feature distributions of the random variables $f_A $ \& $ f_S $ of assistant and the student learner respectively. Here $L_{soft}$ is the BCE Loss between the soft labels generated by the assistant and the student's predictions.

\section{Experiments}
For training and testing our methods, we make use of the CheXpert data-set~\cite{irvin2019chexpert} $^{0}$\footnotetext{https://stanfordmlgroup.github.io/competitions/chexpert/} for the detection of diseases by observing CXRs. The dataset consists of 224316 chest X-ray images  which are labelled individually with 14 different thoracic diseases. A CXR in this dataset can be an indicator of more than one disease simultaneously converting this into a multi-label classification problem.
Initially, we train the teacher network with Adam as an optimizer and a learning rate and weight decay of 1e-3 and 1e-5 respectively. We soften the labels generated from the teacher with a temperature of 20. To train the student and the assistant network we use the Adam optimizer once again with similar hyper-parameters.

We find the overall precision of the student network, taking into account all classes in the 202 images validation data to be \emph{0.952}, while the recall and the F1-score are \emph{0.829} and \emph{0.878} respectively. Table \ref{table:ablation} demonstrates the ablation experiments in this regard. From our experiments in Table \ref{table:ablation}, we find that the assistant network, which in turn is distilled from the teacher network, helps the student learn better (\emph{F-1 score of 0.878}) as compared to direct distillation of the student from the teacher network (\emph{F-1 score of 0.42}). Both these values are quite high compared with training the student network alone (\emph{F-1 score of 0.39})  
\begin{table}[!h]
    \caption{Ablation Experiments}
    \centering
        \begin{tabular*}{\linewidth}{p{70pt}p{40pt}p{40pt}p{40pt}}
            \toprule
            & \textbf{Precision}&\textbf{Recall}&\textbf{F1-Score}\\\midrule
        Teacher alone    &0.968		&0.823		&0.882\\
		Student alone     &0.68      &0.27      &0.39\\
		Student from teacher   &0.69      &0.30      &0.42\\
	    \textbf{Student from assistant}  &\textbf{0.952}	&\textbf{0.829}		&\textbf{0.878}\\
	 \bottomrule\\
	 \label{table:ablation}
	 
        \end{tabular*}
\vspace{-5pt}    
\end{table}

	 

\section{Case Study: COVID-19}
The recent global pandemic of corona-virus has taken the world by storm, resulting in an extreme demand of testing, diagnosis and treatment. The definitive test for COVID-19 being the costly and time consuming process of reverse transcription polymerase chain reaction (RT-PCR), the strain is taking longer to be identified and contained. Inherently, an easily accessible and cheap method of detection is the need of the moment to detect and contain COVID-19. Our proposed method attempts the automated detection of COVID-19 from CXRs to further this cause.


\subsection{Models}
For the detection of Covid19 and Pneumonia from ChestX-ray images, we maintain our training and teaching mechanism similar to the main method with a change in the teacher network backbone. We use the Inception V3 model architecture as the backbone of the \textbf{teacher network}. The Inception V3 is a 42 layer Deep Neural Network architecture consisting of a total of 27 million trainable parameters. We use the Inception v3 model directly on the training dataset and store the extracted feature maps at the end of training.

\subsection{Experiments on case study}

Our dataset for this study consists of 6291 images of CXRs with 4273 images labelled as Pneumonia and 301 images labelled as Covid. We create this dataset from two sources consisting of covid and pneumonia data each \footnote{https://www.kaggle.com/paultimothymooney/chest-xray-pneumonia}
\footnote{https://github.com/ieee8023/covid-chestxray-dataset}. We make use of imbalance prevention methods like under-sampling data from pneumonia class and oversampling data from covid class to prevent data imbalance.
\input{./tables/case_study/table1.tex}
We train the network in a way very similar to that described in our primary method. In here, we use a learning rate of 1e-2, 1e-3, and 1e-3 in the teacher, assistant and the student networks respectively. We use a temperature value of 30 while training the student and the assistant networks. We tabulate the results of our experiments in Table \ref{table:tab2}. 



\section{Conclusion}
In this paper, we demonstrated the possibility of developing a lightweight model that can be trained and used easily in devices with limited hardware and processing capabilities. With fast and accurate detection of diseases from CXRs, we can perform more tests which can aid in early identification and diagnosis diseases, even with limited usage of resources.

\bibliographystyle{IEEEbib}
\bibliography{main}

\end{document}

%% file: fig_eff.tex
\begin{figure}[!ht]
	\centering
		\includegraphics[width=70mm,height=20mm]{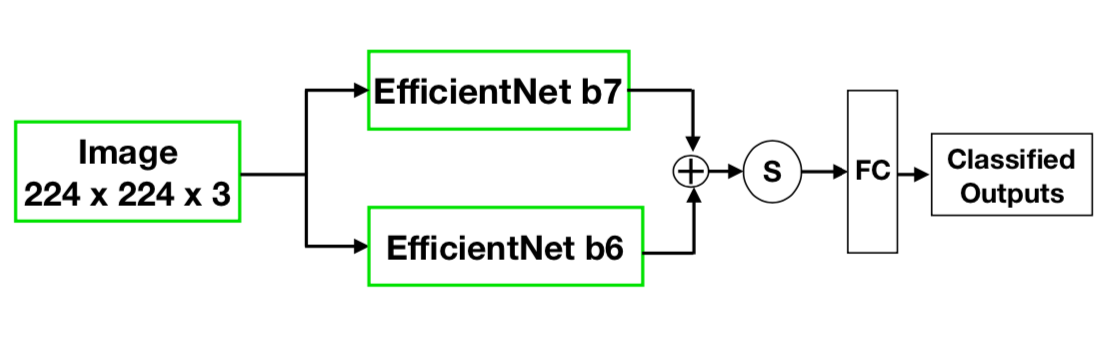}
	\caption{\centering Teacher Network: Efficient-Net ensemble}
	\label{fig:teacher}
\end{figure}

\begin{figure}[!ht]
	\centering
	\subfloat{	
		\includegraphics[width=80mm,height=30mm]{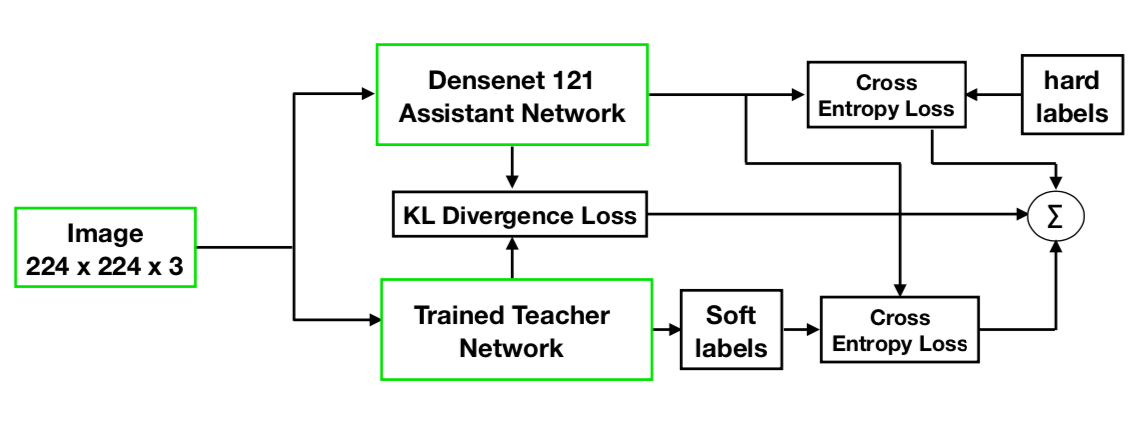}
	}\\\hspace*{-0.5em}\subfloat{
		\includegraphics[width=80mm,height=30mm]{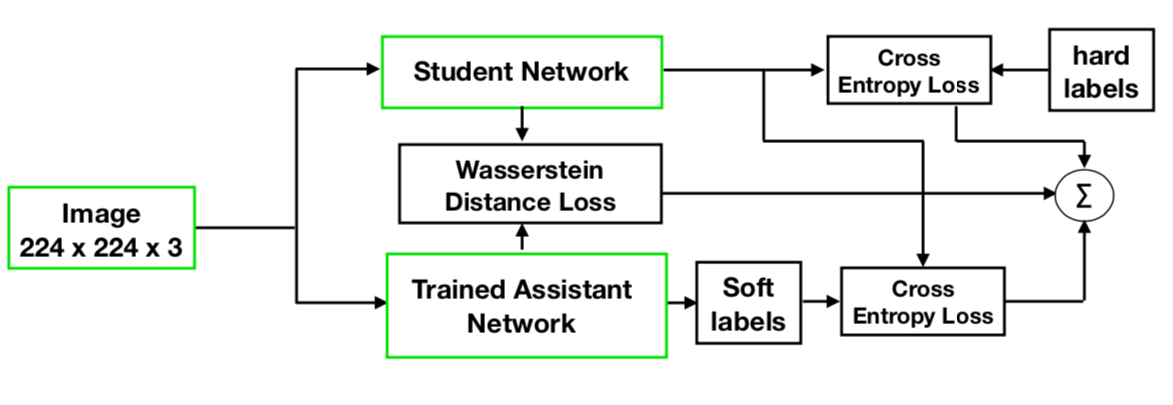}
	}
	
	\caption{\centering(A)Training Assistant Network (B) Training Student network}
	\label{fig:student}
\end{figure}


%% file: tables/case_study/table1.tex
\begin{table}[!h]
    \caption{Classification report for teacher, assistant and student models on Case Study}
    \begin{subtable}{\linewidth}
        \caption{Teacher}
        \begin{tabular*}{\linewidth}{p{60pt}p{40pt}p{40pt}p{40pt}}
            \toprule
            & \textbf{Precision}&\textbf{Recall}&\textbf{F1-Score}\\\midrule
        Normal     &0.9448&0.9778&0.9610\\
		Pneumonia     &0.9917    &0.9789    &0.9853       \\
		Covid     &0.9833    &0.9833    &0.9833        \\
	 Accuracy       &           &        &0.9788      \\
	 Macro avg     &0.9733    &0.9800    &0.9765      \\
	 Weighted avg     &0.9793    &0.9788    &0.9790   \\
	 \bottomrule\\
        \end{tabular*}
    \end{subtable}
    \begin{subtable}{\linewidth}
        \caption{Assistant}
        \begin{tabular*}{\linewidth}{p{60pt}p{40pt}p{40pt}p{40pt}}
            \toprule
            & \textbf{Precision}&\textbf{Recall}&\textbf{F1-Score}\\\midrule
		Normal     &0.948     &0.978     &0.963      \\
		Pneumonia     &0.992    &0.980    &0.986     \\
		Covid     &0.983    &0.983    &0.983     \\
	 Accuracy &          &          &0.980     \\
	 Macro avg&0.974    &0.980    &0.977     \\
	 Weighted avg
	          &0.980    &0.980    &0.980     \\
	 \bottomrule\\
        \end{tabular*}
    \end{subtable}
    \begin{subtable}{\linewidth}
        \caption{Student}
        \begin{tabular*}{\linewidth}{p{70pt}p{40pt}p{40pt}p{40pt}}
            \toprule
            & \textbf{Precision}&\textbf{Recall}&\textbf{F1-Score}\\\midrule
		Normal     &0.941     &0.965     &0.953      \\
		Pneumonia     &0.987    &0.978    &0.982     \\
		Covid     &0.999    &0.999    &0.999     \\
	 Accuracy &          &          &0.976     \\
	 Macro avg&0.976    &0.981    &0.978   \\
	 Weighted avg
	          &0.976    &0.976    &0.976     \\
	 \bottomrule\\
        \end{tabular*}
    \end{subtable}
    \label{table:tab2}
\end{table}
	
	
	
		
	